\documentclass[a4paper,fleqn]{cas-sc}
\usepackage[numbers]{natbib}
\usepackage{multirow}

\usepackage{array}
\usepackage{booktabs}
\usepackage{xcolor}
\usepackage{subcaption} 

\usepackage{amsmath,amssymb,amsfonts}
\usepackage{algorithmic}
\usepackage{graphicx}
\usepackage{textcomp}
\usepackage{wrapfig}
\usepackage{fancyhdr}
\usepackage{float}
\def\tsc#1{\csdef{#1}{\textsc{\lowercase{#1}}\xspace}}
\tsc{WGM}
\tsc{QE}
\tsc{EP}
\tsc{PMS}
\tsc{BEC}
\tsc{DE}

\usepackage{fancyhdr}
\fancypagestyle{first}{%
	\fancyhf{}
	\fancyfoot[C]{\thepage}

}
\fancypagestyle{plain}{%
	\fancyhf{}
	\fancyfoot[C]{\thepage}

}
\begin{document}
\let\WriteBookmarks\relax
\def\floatpagepagefraction{1}
\def\textpagefraction{.001}
\shorttitle{}
\shortauthors{}

\title [mode = title]{D-Optimality-Guided Reinforcement Learning for Efficient Open-Loop Calibration of a 3-DOF Ankle Rehabilitation Robot}

\author[1]{Qifan Hu}

\author[2]{Branko Celler}
 
\author[1]{Weidong Mu}
\cormark[1]
 
\author[3]{Steven W. Su}
\cormark[1]

\affiliation[1]{organization={Affiliated Provincial Hospital, Shandong First Medical University},
	addressline={324 Jing Wu Road}, 
	city={Jinan},
	postcode={250021}, 
	state={Shandong},
	country={China}}

\affiliation[2]{organization={Faculty of Engineering, University of New South Wales},
	postcode={NSW 2052}, 
	country={Australia}}
	
\affiliation[3]{organization={Faculty of Engineering and IT, University of Technology Sydney},
	postcode={NSW 2007}, 
	country={Australia}}

\begin{abstract}
Accurate alignment of multi-degree-of-freedom rehabilitation robots is essential for safe and effective patient training. This paper proposes a two-stage calibration framework for a self-designed three-degree-of-freedom (3-DOF) ankle rehabilitation robot. First, a Kronecker-product-based open-loop calibration method is developed to cast the input–output alignment into a linear parameter identification problem, which in turn defines the associated experimental design objective through the resulting information matrix. Building on this formulation, calibration posture selection is posed as a combinatorial design-of-experiments problem guided by a D-optimality criterion, i.e., selecting a small subset of postures that maximises the determinant of the information matrix.

To enable practical selection under constraints, a Proximal Policy Optimization (PPO) agent is trained in simulation to choose 4 informative postures from a candidate set of 50. Across simulation and real-robot evaluations, the learned policy consistently yields substantially more informative posture combinations than random selection: the mean determinant of the information matrix achieved by PPO is reported to be more than two orders of magnitude higher (>100×) with reduced variance. In addition, real-world results indicate that a parameter vector identified from only four D-optimality-guided postures provides stronger cross-episode prediction consistency than estimates obtained from a larger but unstructured set of 50 postures. The proposed framework therefore improves calibration efficiency while maintaining robust parameter estimation, offering practical guidance for high-precision alignment of multi-DOF rehabilitation robots.

\end{abstract}


\maketitle

\section{Introduction}
Rehabilitation therapy plays a critical role in restoring lower-limb function. Previous studies have demonstrated that repetitive, high-intensity, and task-oriented rehabilitation is essential for promoting neural plasticity and functional recovery \cite{cramer2011harnessing, kim2024using}. Conventional gait training (CGT) typically relies on therapist assistance: patients who are able to walk use walkers for supported gait training, whereas patients with limited mobility perform pre-walking exercises such as trunk balance training, sitting and standing balance exercises, and isolated lower-limb movements \cite{talaty2023feasibility}. However, traditional rehabilitation training methods suffer from two major limitations: the lack of feedback on human–machine interaction and the absence of intelligent data-driven training feedback \cite{wang2023modeling}.

Robotic-assisted rehabilitation has emerged as a promising solution to overcome these limitations by automating training processes and providing repetitive, intensive, and task-oriented exercises \cite{morone2017robot, nizamis2021converging, poli2013robotic, laszlo2023feasibility}. In particular, repetitive ankle movements have been shown to effectively improve joint range of motion and neuromuscular coordination. Robotic devices further enable therapists to supervise multiple patients simultaneously and support more complex rehabilitation tasks, thereby enhancing rehabilitation efficiency and promoting neural plasticity \cite{stanton2011biofeedback, diego2024devices}.

In our preliminary work, we developed and prototyped a compact 3-DOF ankle rehabilitation robot that supports both single-DOF and coupled multi-DOF motions.
Beyond mechanical design, effective control remains a central issue in rehabilitation robotics \cite{rodriguez2021systematic, diaz2022human}. Existing approaches typically guide the ankle--foot complex along reference trajectories and can be broadly categorised into trajectory-tracking control and assist-as-needed (AAN) strategies \cite{hussain2021robot, abarca2023review}. For multi-DOF devices, accurate posture alignment is a prerequisite for both categories, because misalignment directly propagates into tracking errors, inappropriate assistance, and potentially unsafe interaction forces.

For the proposed 3-DOF ankle rehabilitation robot, posture alignment constitutes a multi-input multi-output (MIMO) problem \cite{gao2010sensor}. In practice, discrepancies between commanded motor inputs and the actual end-effector postures are inevitable due to mechanical tolerances, transmission backlash, and assembly errors. Such misalignment degrades control accuracy, compromises rehabilitation performance, and may introduce safety risks for patients.

A straightforward approach is to construct a closed-loop system by mounting posture sensors on the end-effector to provide real-time feedback for calibration and control \cite{10845160}. Common sensing technologies include vision-based systems \cite{s25020498}, Global Positioning System (GPS) receivers \cite{kumar2002evolution}, ultrasonic sensors \cite{qiu2022review}, and inertial measurement units (IMUs) integrating accelerometers, gyroscopes, and magnetometers \cite{zhang2023recent, NAZARAHARI202167, KHOR2025}. Although IMUs can estimate full attitude, their performance is often limited by gyroscope drift and magnetometer sensitivity to electromagnetic interference \cite{Han_2025}, particularly in indoor environments \cite{renaudin2010complete, guo2008soft, yu2022statistical}. GPS-assisted heading correction similarly suffers from reduced accuracy in indoor or signal-degraded scenarios \cite{chang2023review, WANG2025101977}. More importantly for rehabilitation robots, the strict timing constraints of the motion--measurement loop make reliable real-time attitude sensing, synchronisation, and filtering challenging under dynamic conditions.

To mitigate the impact of sensing limitations on control accuracy, an open-loop calibration strategy can be adopted. Rather than relying on continuous high-dynamic posture feedback during operation, a one-time calibration experiment is performed using steady-state high-precision sensing to collect paired input--output posture data. Open-loop calibration exploits steady-state measurements at selected postures, thereby bypassing high-dynamic sensing requirements and improving measurement accuracy and repeatability. An input-output mapping model is identified from the calibration dataset and embedded into the controller, enabling accurate open-loop posture control without real-time feedback during subsequent use \cite{papafotis2020multiple, gheorghe2023disentangling}.

However, calibration accuracy must be balanced against calibration time, experimental burden, and safety. In our system, collecting a candidate set of 50 postures per session is feasible, but utilizing all of them for identification is inefficient. The key challenge is to select a much smaller subset of postures that remains sufficiently informative for accurate and robust estimation of the calibration parameters.

In this paper, we propose a \emph{Kronecker-product-based open-loop posture calibration} method for a custom 3-DOF ankle rehabilitation robot. The resulting calibration model requires the estimation of \emph{twelve parameters}. The Kronecker-based formulation implies that, under ideal rank conditions, only a small number of distinct postures is theoretically sufficient for identifiability. This motivates a principled experimental design problem: how to minimise the number of calibration postures while preserving maximal information for parameter estimation.

To address this, we built a simulation model based on the robot's SolidWorks design to generate calibration data and support offline optimisation. We then developed a simulation-guided framework that integrates reinforcement learning with a \emph{$D$-optimality-inspired} experimental design objective, where the informativeness of a selected posture subset is quantified by the log-determinant of the corresponding information matrix. Specifically, a Proximal Policy Optimisation (PPO) agent is trained to sequentially select a small yet informative subset of postures---as few as four, consistent with the minimum identifiability requirement---from each candidate set of 50. Extensive simulation studies and practical experiments demonstrate that the learned policy consistently identifies informative posture combinations, significantly reducing calibration effort while maintaining high-quality parameter identification. More broadly, the proposed framework provides a general solution for \emph{simulation-guided experimental point selection} in multi-DOF robotic systems under strict measurement budgets.

The main contributions of this work are as follows:
\begin{enumerate}
	\item A Kronecker-product-based open-loop posture calibration method for a self-designed 3-DOF ankle rehabilitation robot, enabling estimation of a 12-parameter calibration model from one-time high-precision calibration data without relying on real-time feedback during operation.
	
	\item A discrete posture-subset selection formulation induced by the proposed calibration model, together with a learning-based solver that combines a $D$-optimality information criterion (log-determinant) with a PPO policy to obtain \emph{approximate} information-maximising posture sets under strict calibration budgets.
	
	\item Comprehensive simulation and real-world validation on a compact 3-DOF ankle rehabilitation robot, demonstrating that simulation-guided, information-maximising posture selection can substantially reduce calibration effort while maintaining robust parameter estimation.
\end{enumerate}

To the best of our knowledge, while reinforcement-learning-based sequential experimental design has been explored in general settings, its use as a practical tool for $D$-optimality-inspired posture subset selection tailored to \emph{open-loop} calibration of multi-DOF rehabilitation robots has been rarely studied in the literature.

\section{Methodology}
In this section, a Kronecker-product-based open-loop posture calibration method is introduced for system parameter identification. Furthermore, to design informative experiments that maximize parameter identifiability and estimation accuracy while minimizing experimental cost and effort, an optimized experimental posture selection method is proposed, which integrates D-optimality-guided experimental design with reinforcement learning.

The proposed PPO-based policy optimization framework provides a general, simulation-guided solution for experimental point optimization, enabling efficient and robust experimental design using only system input information, without reliance on explicit analytical model structures.

\subsection{System Modeling and Parameter Identification}
The self-designed 3-DOF ankle rehabilitation robot is modeled as a multiple-input multiple-output (MIMO) system. Based on this model, an open-loop calibration-based parameter identification approach is developed to improve alignment accuracy. Subsequently, the proposed calibration algorithm is implemented for robot alignment.

\subsubsection{MIMO System Formulation}
This system has three inputs: the pitch angle control input $u_x$, yaw angle control input $u_y$, and roll angle control input $u_z$, which together form the input state vector $U=[u_x\,\,\,u_y\,\,\,u_z]^T$. The corresponding output state vector is $Y=[y_x\,\,\,y_y\,\,\,y_z]^T$. After conducting $N$ alignment experiments, we obtained $N$ sets of input state vectors and $N$ corresponding output state vectors:

\begin{center}

$U_1=[u_{x_1}\,\,\,u_{y_1}\,\,\,u_{z_1}]^T$,
 $U_2=[u_{x_2}\,\,\,u_{y_2}\,\,\,u_{z_2}]^T$,
 \dots , 
 $U_N=[u_{x_N}\,\,\,u_{y_N}\,\,\,u_{z_N}]^T$ ,

\end{center}
and 
\begin{center}

$Y_1=[y_{x_1}\,\,\,y_{y_1}\,\,\,y_{z_1}]^T$,
$Y_2=[y_{x_2}\,\,\,y_{y_2}\,\,\,y_{z_2}]^T$,
\dots ,   
$Y_N=[y_{x_N}\,\,\,y_{y_N}\,\,\,y_{z_N}]^T$ .
\end{center}

The linear relationship between the input posture and the output posture can be expressed as follows:

\begin{equation} \label{eq1}
Y
=
\begin{bmatrix}
a_{11} & a_{12} & a_{13} \\
a_{21} & a_{22} & a_{23} \\
a_{31} & a_{32} & a_{33}
\end{bmatrix}
U
+
\begin{bmatrix}
b_x \\
b_y \\
b_z
\end{bmatrix} ,
\end{equation}

\begin{equation} \label{eq10}
	Y=
	X_A
	U
	+
	X_B,
\end{equation}
where the parameter matrices are defined as follows:
\begin{equation}\label{eq5}
	X_A=
	\begin{bmatrix}
		a_{11}&a_{12}&a_{13}\\
		a_{21}&a_{22}&a_{23}\\
		a_{31}&a_{32}&a_{33}
	\end{bmatrix},
\end{equation}
\begin{equation}\label{eq6}
	X_B=
	\begin{bmatrix}
		b_x&
		b_y&
		b_z
	\end{bmatrix}^T.
\end{equation}

It should be noted that in Equation (\ref{eq5}), the element $a_{11}$ represents the scaling factor of the pitch angle for the end-effector of a three-axis robot, the element $a_{22}$ represents the scaling factor of the yaw angle and $a_{33}$ represents the scaling factor of the roll angle. In Equation (\ref{eq6}), the elements $b_x$, $b_y$, and $b_z$ represent the biases of the three-axis robot's final pitch, yaw, and roll axes, respectively. 

\subsubsection{Kronecker-Based Open-Loop Parameter Identification}

Based on the above equations, it is clear that in the proposed algorithm, we need to set $12$ calibration parameters. In order to identify the parameters using the well-developed standard Least Square approach, we have to transfer the parameters into a single vector. In this section, we apply the Kronecker product approach for the first item  of the right side of Equation (\ref{eq10}), using a single vector $vec(X_A)=[a_{11}\,\,\, a_{12}\,\,\,a_{13}\,\,\,a_{21}\,\,\,a_{22}\,\,\,a_{23}\,\,\,a_{31}\,\,\,a_{32}\,\,\,a_{33}]^T$ in the following format:

 \begin{equation} \label{eq8}
vec(Y)=(I_3\otimes
 	U^T)
 	vec(X_A)+vec(X_B).
   \end{equation} 
	That is,
\begin{equation}
	Y = \begin{bmatrix}
		U^T& 0 & 0\\
		0 & U^T & 0\\
		0 & 0 & U^T\\
	\end{bmatrix} vec(X_A) + X_B .
\end{equation}

Denote $X=[ vec(X_A)^T \,\,\,\vert\,\,\, X_B^T]^T$, we have  
\begin{equation} \label{eq9}
	Y=
	\begin{bmatrix}
		U^T & 0 & 0 & \vert & 1 & 0 & 0\\
		0 & U^T & 0 & \vert & 0 & 1 & 0\\
		0 & 0 & U^T & \vert & 0 & 0 & 1
	\end{bmatrix}X .
\end{equation}

When considering the total $N$ sets of experimental data, Equation (\ref{eq9}) can be augmented in the following form:
\begin{equation} \label{eq2}
\begin{bmatrix}
y_{x_1} \\
y_{y_1} \\
y_{z_1}\\
\vdots \\
y_{x_N} \\
y_{y_N} \\
y_{z_N}\\
\end{bmatrix}
=
\begin{bmatrix}
U_1^T & 0 & 0 & 1 & 0 & 0\\
0 & U_1^T & 0 & 0 & 1 & 0\\
0 & 0 & U_1^T & 0 & 0 & 1\\
\vdots & \vdots & \vdots & \vdots & \vdots & \vdots\\
U_N^T & 0 & 0 & 1 & 0 & 0\\
0 & U_N^T & 0 & 0 & 1 & 0\\
0 & 0 & U_N^T & 0 & 0 & 1
\end{bmatrix}
\begin{bmatrix}
a_{11}\,\,\,a_{12}\,\,\,a_{13}\,\,\,a_{21}\,\,\,a_{22}\,\,\,
a_{23}\,\,\,a_{31}\,\,\,a_{32}\,\,\,a_{33}\,\,\,b_x\,\,\,b_y\,\,\,b_z
\end{bmatrix}^T .
\end{equation}

The compact form of the above equation can be expressed as follows:
\begin{equation} \label{eq3}
    \bar Y=AX+\epsilon, 
\end{equation}
where $\bar Y= [y_{x_1} \,\,\, y_{y_1} \,\,\,y_{z_1} \,\,\,\cdots \,\,\,y_{x_N}\,\,\, y_{y_N} \,\,\, y_{z_N}]^T$.

In Equation (\ref{eq3}), when considering measurement noise, a vector $\epsilon$ is added, whose elements are white noises with zero mean. 

According to the classical least-square approach, the parameter estimate is given by:
\begin{equation} \label{eq4}
    X=(A^T A)^{-1} A^T  \bar Y.
\end{equation}

It should be noted that although Equation (\ref{eq5}) and (\ref{eq6}) indicate that a total of $12$ parameters need to be identified, Equation (\ref{eq2}), obtained through the Kronecker product–based reformulation, reveals that only four distinct experimental points are theoretically sufficient to identify all parameters.

The accuracy and robustness of system parameter identification are strongly dependent on the quality of experimental data. Although the proposed open-loop parameter identification method enables effective estimation of the unknown model parameters, the identifiability and estimation accuracy are inherently influenced by the selection of experimental operating points. Inappropriate or poorly distributed experimental points may lead to ill-conditioned regression matrices and degraded identification performance.

In addition to estimation accuracy considerations, practical constraints often limit the number of feasible experimental points in robotic systems. Excessive data collection may increase mechanical wear, risk safety violations, and prolong experimental time. Consequently, an optimized experimental point selection strategy is required to maximize identification performance using a limited set of experimental data.

\begin{figure}[h!]
	\centering
	\includegraphics[width=\textwidth]{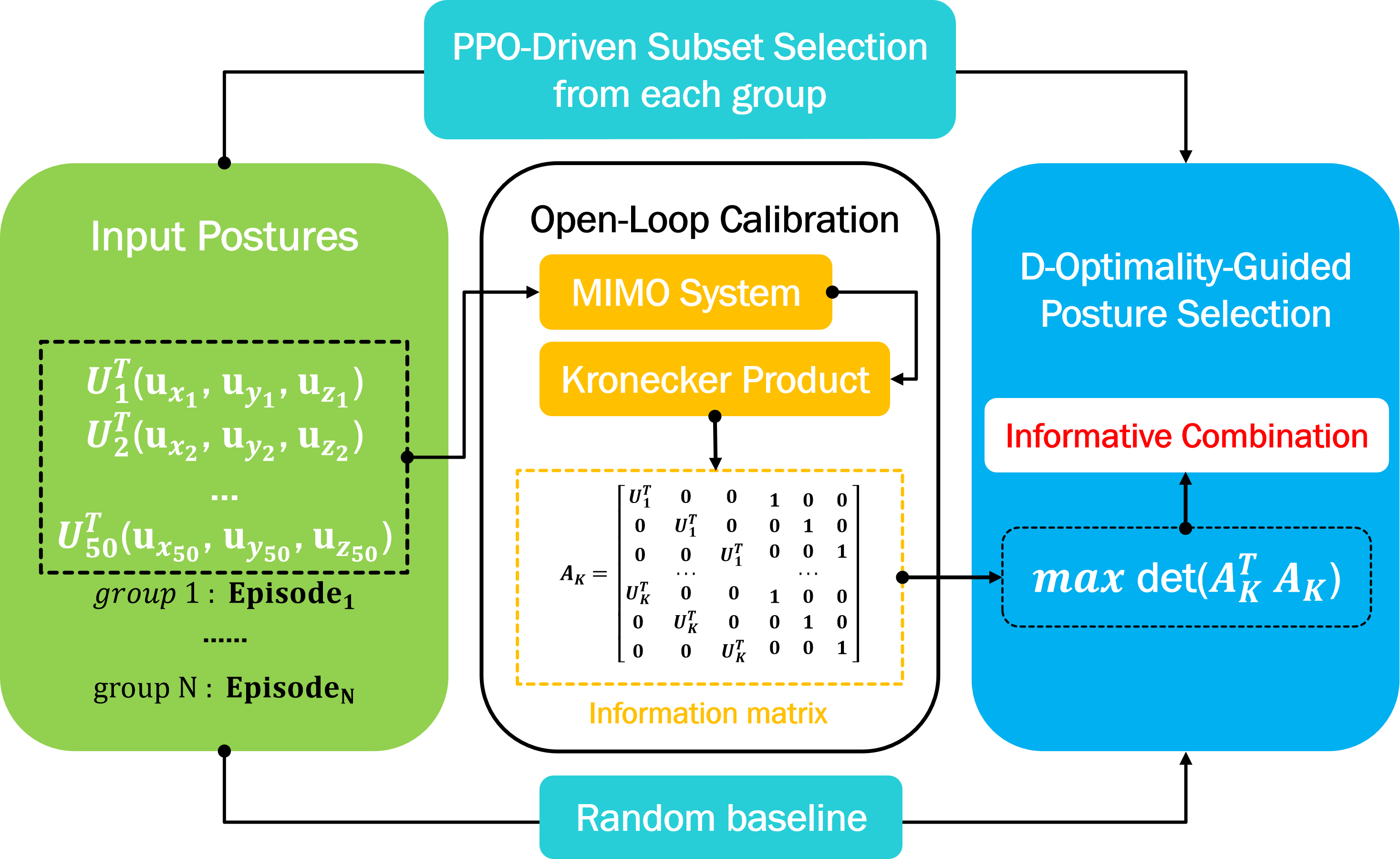}
	\caption{Formulation of a D-Optimality-Guided Posture Selection Problem.}
	\label{method}
\end{figure}
\subsection{Experimental Point Optimization via Design of Experiments}

In this part, a general and effective experimental point optimization strategy is presented. The proposed approach leverages simulation to guide the selection of experimental points in real-world settings and performs optimization based solely on the system inputs, without requiring additional system output feedback during the optimization process. Moreover, the explicit system model is required only during the training stage, but not during the application stage, which enables a general and flexible experimental point optimization framework. 
In summary, the experimental point optimization problem can be interpreted as a design of experiments (DoE) problem aimed at maximizing parameter identifiability.
This design facilitates practical experimental implementation under constrained conditions and improves the accuracy and robustness of subsequent parameter estimation. 

\subsubsection{Problem Formulation and Design Objectives}

To reformulate the experimental point selection problem as a DoE problem aimed at improving parameter identifiability, we consider selecting a subset of experimental postures from a finite candidate set.

Let $\mathcal{P}$ denote the set of all feasible postures. 
At the $i$-th episode, a candidate posture set
$\mathcal{P}_{\mathrm{cand}}^{(i)} \subseteq \mathcal{P}$ with cardinality $M$ is given.
From this set, a subset
$\mathcal{P}_{\mathrm{opt}}^{(i)} \subseteq \mathcal{P}_{\mathrm{cand}}^{(i)}$
containing $K$ postures is selected, where $M$ and $K$ are positive integers satisfying $4 \le K < M$.

For a given posture subset $\mathcal{P}_{\mathrm{opt}}^{(i)}$,
the corresponding information matrix is constructed as
\begin{equation}\label{eq_S}
	\mathbf{S}\big(\mathcal{P}_{\mathrm{opt}}^{(i)}\big)
	= A^\top\big(\mathcal{P}_{\mathrm{opt}}^{(i)}\big)
	A\big(\mathcal{P}_{\mathrm{opt}}^{(i)}\big),
\end{equation}

The experimental design problem is then formulated under the D-optimality criterion as
\begin{equation} \label{eq_op}
	\begin{aligned}
		\max_{\mathcal{P}_{\mathrm{opt}}^{(i)} \subseteq \mathcal{P}_{\mathrm{cand}}^{(i)}} 
		& \quad \det \Big( \mathbf{S}\big(\mathcal{P}_{\mathrm{opt}}^{(i)}\big) \Big) \\
		\text{s.t.} 
		& \quad \left| \mathcal{P}_{\mathrm{opt}}^{(i)} \right| = K .
	\end{aligned}
\end{equation}

From a practical perspective, acquiring a moderate number of experimental postures in a single calibration session helps reduce mechanical wear, ensure operational safety, and limit the overall experimental duration. Accordingly, the number of experimental postures collected in each acquisition is fixed at 
$M = 50$.

From a theoretical standpoint, at least four experimental postures are required to identify the twelve unknown calibration parameters. Therefore, the optimization objective is to select a subset of 
$K = 4$ postures from each candidate set of 50 postures that maximizes the information content of the collected data while maintaining practical feasibility.

In summary, under the considered problem setting, the experimental point selection problem aims to identify four informative postures from each candidate set of 50 available postures by maximizing the determinant of the corresponding information matrix.

\subsubsection{D-Optimality-Guided Optimization Criterion}

To quantitatively evaluate the informativeness of experimental points for parameter identification, an optimization criterion guided by the D-optimality principle is adopted. From a parameter estimation perspective, the accuracy of parameter identification is directly related to the conditioning and information content of the regression matrix. Under standard assumptions, this relationship can be characterized by the information matrix, whose inverse provides a lower bound on the parameter covariance. Maximizing the determinant of the information matrix, or equivalently its logarithmic determinant, enhances parameter identifiability and reduces estimation uncertainty, which forms the theoretical basis of the D-optimal design criterion.

In this study, the D-optimality criterion is employed as a design objective to guide experimental point selection. Notably, the simulation data are generated by constraining the spatial motion angles of the three rotational axes of the ankle rehabilitation robot within predefined ranges, which can be specified according to different mechanical structures and the required range of ankle motion. \emph{This input-driven formulation enables the evaluation of information content during offline training using a structured regression model, while no explicit system model is required during the deployment stage,} thereby providing a general and flexible optimization criterion suitable for simulation-guided experimental point optimization.

Based on the above criterion, the logarithmic determinant of the information matrix is employed as the optimization objective and subsequently incorporated as the reward signal in a reinforcement learning framework. The detailed reinforcement learning formulation and training procedure are presented in the following subsection.

It should be noted that the structured system model is only utilized to compute the reward signal during the offline reinforcement learning training process. Once training is completed, the learned policy directly outputs optimized experimental points based solely on system input constraints, without requiring access to the system model or real-time output feedback during practical deployment.

\subsubsection{Simulation-Guided Reinforcement Learning Framework}
To efficiently optimize experimental points under practical constraints, a simulation-guided reinforcement learning (RL) framework is developed. The key idea is to decouple experimental point optimization from real-time data collection by performing policy learning entirely in a simulation environment, while deploying the optimized experimental points in real-world experiments for parameter identification. This simulation-to-real separation significantly reduces experimental cost and avoids unnecessary hardware wear.

Within the proposed reinforcement learning framework, the reward function is constructed based on the information matrix evaluated using a structured system model. Specifically, for a given set of selected experimental points, the corresponding regression matrix is constructed according to the predefined model structure, and the information matrix is subsequently computed. The logarithmic determinant of the information matrix is then used to quantitatively evaluate the overall informativeness of the selected experimental points.

To reflect practical experimental constraints and encourage the selection of complementary experimental points, the reward is designed to be sparsely evaluated. In particular, the reinforcement learning agent receives a reward only after every four experimental points are selected, which corresponds to the minimum batch size required to form a full-rank regression matrix under the adopted parameterization, rather than after each individual action. This delayed reward mechanism allows the information contribution of a group of experimental points to be jointly assessed, which is consistent with the batch construction of the regression matrix based on the Kronecker product formulation.

Under this formulation, the agent performs a sequence of actions to generate experimental points, while the environment accumulates the corresponding input configurations. Once a predefined batch size is reached, the information matrix is computed based on the accumulated experimental points, and the resulting log-determinant value is provided as the reward signal. This design facilitates stable policy learning and encourages consistency between the reinforcement learning objective and the D-optimality criterion employed for experimental point selection.

\subsubsection{Policy Optimization Using Proximal Policy Optimization}

\begin{figure}[h!]
	\centering
	\includegraphics[width=0.6\textwidth]{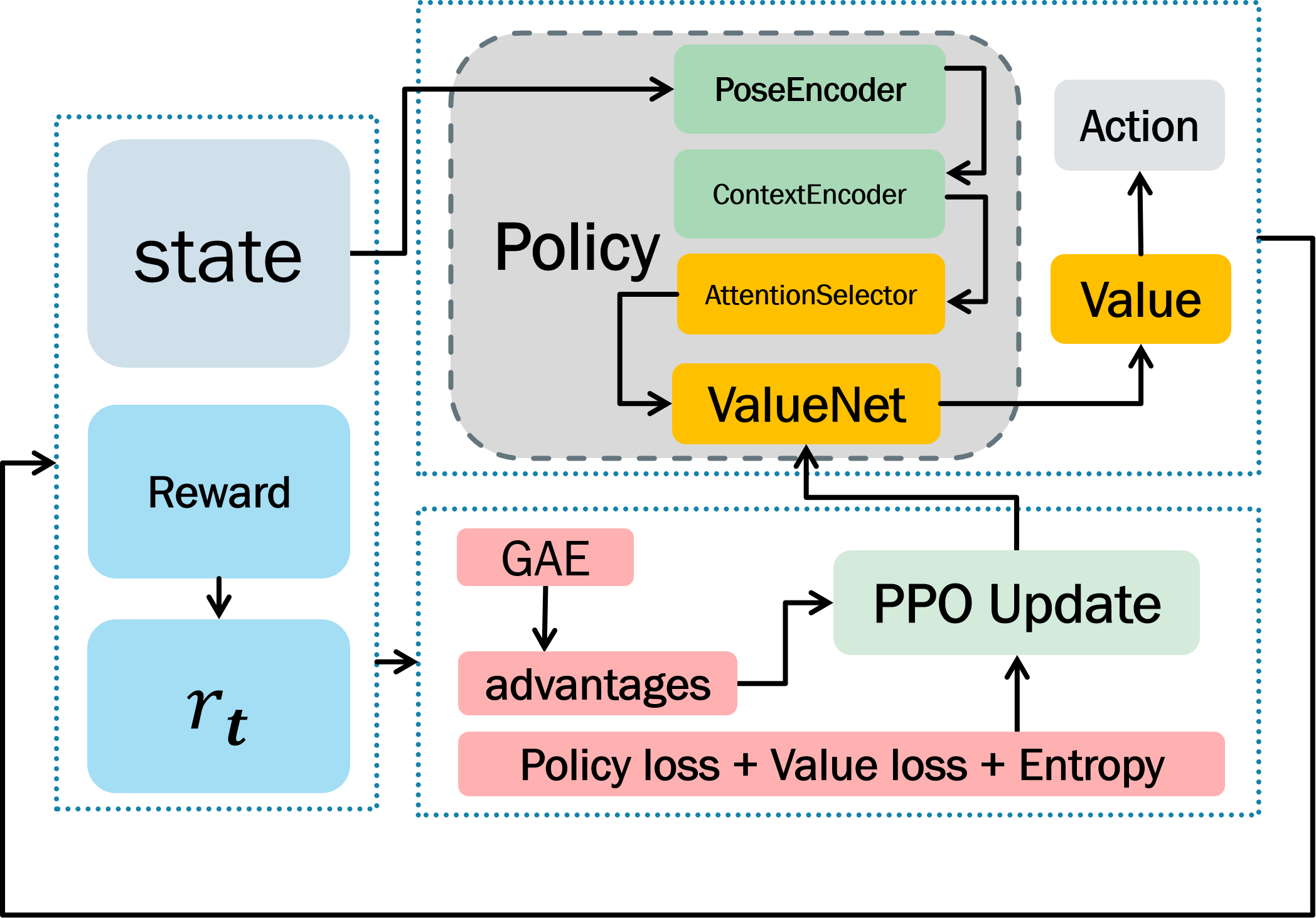}
	\caption{PPO-based RL Frame Workflow.}
	\label{fram}
\end{figure}

The policy is parameterized using an actor–critic architecture, in which a policy network (actor) selects experimental points and a value network (critic) estimates the state value to reduce the variance of policy gradient updates. An attention-based policy network is employed to handle the combinatorial nature of experimental point selection: candidate experimental points are individually encoded, while the current optimization context—including statistics of previously selected points, optimization progress, and matrix-related information—is processed by a shared context encoder. An attention mechanism then computes selection scores for all feasible experimental points by evaluating their relevance to the encoded context. The policy outputs a categorical distribution over the available experimental points, from which actions are sampled during training, while the value network produces a scalar state-value estimate. Policy updates are performed using Proximal Policy Optimization with a clipped surrogate objective, combined with a mean-squared error loss for value function approximation and an entropy regularization term to encourage exploration. The entire optimization process is conducted in simulation using only system input information, and the trained policy can be directly deployed for real-world experimental point selection without requiring additional output feedback or explicit analytical model structures.

\section{Simulation and Analysis}
In this section, two simulation studies are conducted for different purposes.
The first simulation is designed to preliminarily verify the effectiveness of the proposed Kronecker-product-based parameter identification model.
The second simulation focuses on the training and analysis of the reinforcement-learning-based experimental design method.
Specifically, the PPO policy network is trained in a simulation environment using a reward function primarily designed to promote higher information content, as quantified by the determinant of the information matrix. Standard PPO exploration strategies are adopted during training.
Subsequently, the trained PPO policy and a random selection baseline are evaluated on newly generated simulation datasets. The determinant of the information matrix is recorded as the primary evaluation metric, and both the mean and standard deviation are computed to assess performance and stability.
\subsection{Kronecker-Based Parameter Identification Validation}

In this part, a simulation environment for the 3-DOF ankle rehabilitation robot is first developed using the SimMechanics toolbox in MATLAB/Simulink, based on the mechanical structure designed in SolidWorks. The simulation environment is then employed to set system parameters and use simulation $Dataset_0$ (see Table \ref{tab3}) for algorithm validation.
\subsubsection{Simulation Setup and Data collection}

\begin{figure}[h!]
	\centering
	\includegraphics[width=0.8\textwidth]{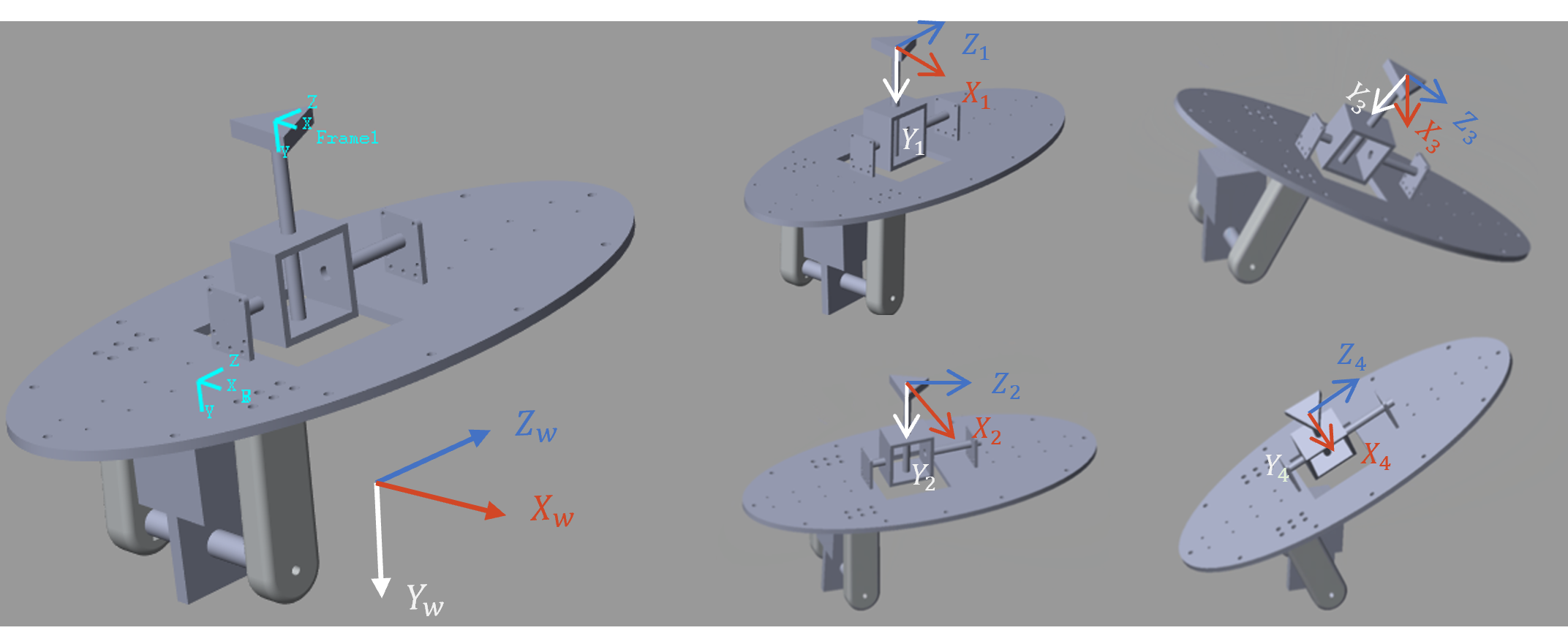}
	\caption{End-effector attitude change relative to the world coordinate system.}
	\label{fig_3}
\end{figure}

Before conducting simulations, it is essential to establish a world coordinate system and set up a reference coordinate system at the end-effector of the three-axis ankle rehabilitation robot. As shown in Fig.~\ref{fig_3}, the world coordinate system \((X_w, Y_w, Z_w)\) remains fixed, while the reference coordinate system \((Frame1)\) initially coincides with the world coordinate system at the base position \((B)\) and changes synchronously with the orientation of the end-effector. The relative attitude change of the reference coordinate system \((Frame1)\) with respect to the base coordinate system \((B)\) accurately reflects the motion of the end-effector from its initial position.

For validation purposes, artificial scaling factors and bias parameters are imposed on each system axis in the simulation environment. The detailed parameter settings are provided in Table \ref{tab1}.
\begin{table}[h!]
	\centering
	\caption{Set values}
	\label{tab1}
	\resizebox{0.5\linewidth}{!}{%
		\begin{tabular}{lccc}
			\toprule
			& Pitch & Roll & Yaw \\ 
			\midrule
			Scaling factor  & $0.43$ & $0.87$ & $0.71$ \\ 
			Bias & $3.1$ & $2.41$ & $-5.8$ \\ 
			\bottomrule
		\end{tabular}
	}
\end{table}

Data collection is conducted in the simulation environment under open-loop control. Provide the system with three independent random input signals: one for controlling the pitch angle, one for controlling the roll angle, and one for controlling the yaw angle. Each signal should vary randomly over time to test the system's response to dynamic, unpredictable inputs in all three rotational axes. The robot model is driven through a sequence of predefined postures, and the system input signals $U$ together with the corresponding output responses $Y$ are sampled at a fixed frequency of $1$ Hz. The simulation duration for each group is $50$ s, resulting in a total of $200$ simulation groups, which is shown in Table \ref*{tab3} . The resulting $Dataset_0$ are used for subsequent parameter identification and performance evaluation.

\subsubsection{Simulation Results and Analysis}
\label{Simulation Results and Analysis}

\begin{figure}[h!]
	\centering
	\includegraphics[width=0.5\textwidth]{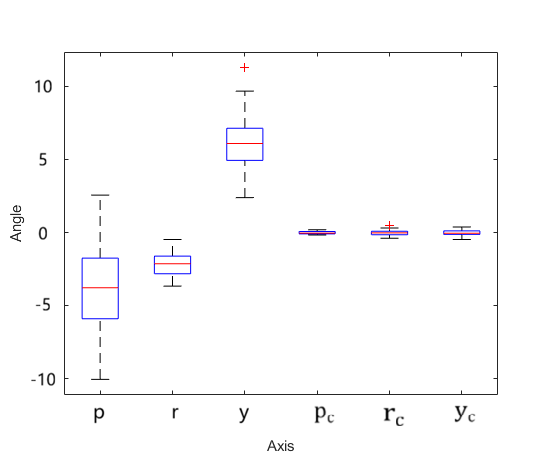}
	\caption{Distribution of input-output errors before and after open-loop calibration in simulation experiments.}
	\label{fig:0}
\end{figure}

Figure~\ref{fig:0} illustrates the boxplot comparison of input–output errors for each axis before and after calibration based on the simulation $Dataset_0$ collected before. Prior to calibration, significant systematic biases and large error dispersion can be observed, particularly along the yaw axis. After applying the proposed parameter identification algorithm according to Equation (\ref{eq4}), the median errors of all axes are driven close to zero, and the interquartile ranges are substantially reduced. These results indicate that the proposed method effectively compensates for axis-wise scaling and bias errors, leading to improved accuracy and robustness.

\subsection{Experimental Point Optimization validation}
This part presents the training and evaluation of a PPO-based experimental point optimization model and demonstrates the practical performance of the proposed experimental design approach on both simulated and real-time data. Specifically, this study proposes a RL–based posture selection method that incorporates the D-optimality criterion into a PPO framework to reduce the complexity of the DoE procedure. The experimental evaluation aims to systematically assess the optimization performance of the proposed method in comparison with a random selection baseline, examine its generalization capability and stability in a simulated environments.

To ensure fairness and reliability of the experimental evaluation, the training and testing phases were performed on separate datasets, and the random-selection baseline was executed multiple times, with its results averaged to obtain stable performance estimate. In addition, all strategies performed the same number of pose selections ($4$ out of $50$ postures) to ensure a consistent evaluation setting.
\begin{figure}[h!]
	\centering
	
	\includegraphics[width=0.8\textwidth]{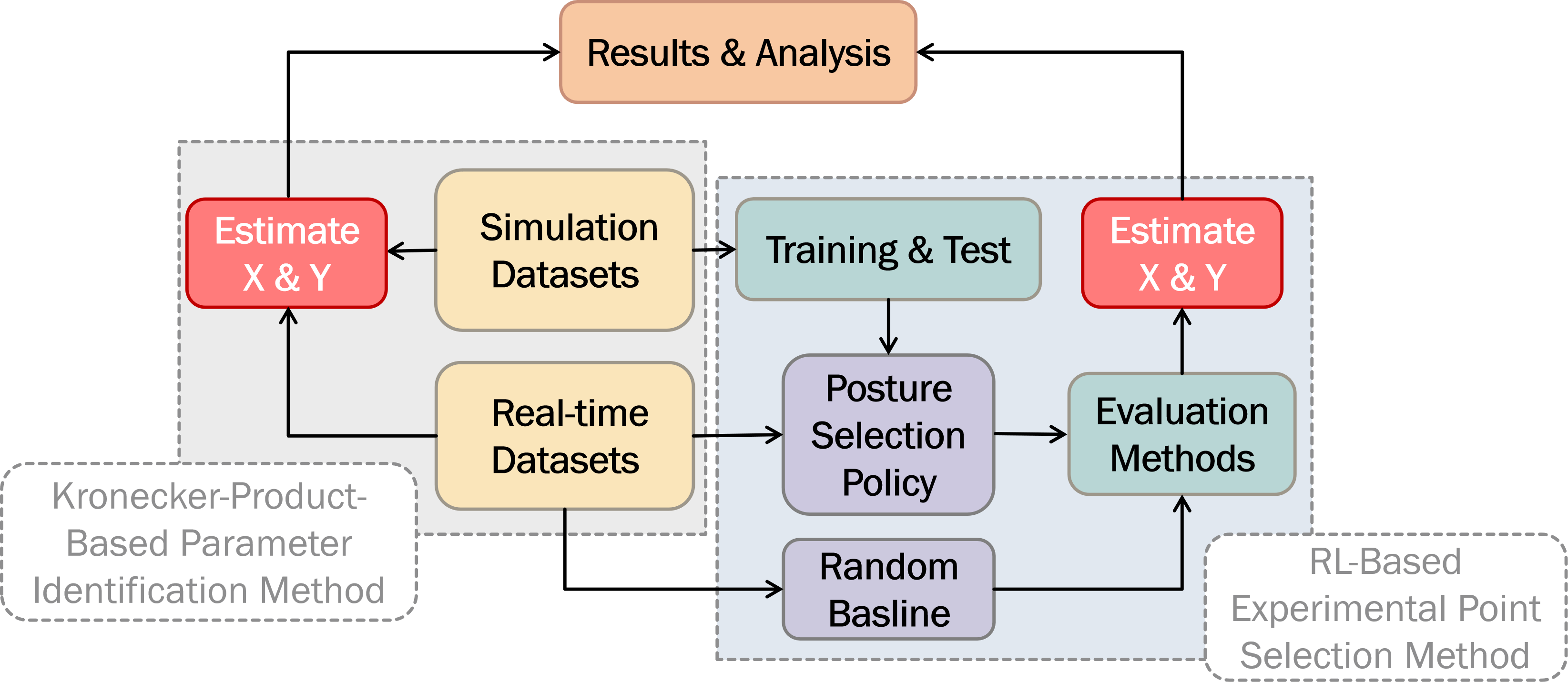}
	
	\caption{Experimental Procedure.}
	\label{ex_pro}
\end{figure}

\begin{table}[h!]
	\centering
	\caption{Datasets}
	\label{tab3}
	\resizebox{0.85\linewidth}{!}{
		\begin{tabular}{l l c c c c}
			\toprule
			\textbf{No.} & \textbf{Dataset Type} & \textbf{Episode (N)} & \textbf{Dim} & \textbf{Collect-num} & \textbf{Select-num} \\ 
			\midrule
			0 & Simulation     & $200$ & $6$ & $50 \times N$ & None \\ 
			1 & Simulation     & $200000$ & $3$ & $50 \times N$ & $4 \times N$ \\ 
			2 & Simulation      & $1000$   & $3$ & $50 \times N$ & $4 \times N$ \\ 
			3 & Simulation  & $100$    & $6$ & $50 \times N$ & $4 \times N$ \\ 
			4 & Real-time        & $100$    & $6$ & $50 \times N$ & $4 \times N$ \\ 
			5 & Real-time      & $1$      & $6$ & $50$          & $4$ \\ 
			\bottomrule
		\end{tabular}
	}
\end{table}

\subsubsection{Simulation Setup}
All experiments were conducted using Python 3.8 and PyTorch framework, which served as the platform for implementing and training the PPO-based reinforcement learning network.

\subsubsection{Simulation Datasets}
Simulation datasets are generated within fixed angular ranges for each axis to ensure that all inputs remain within the admissible input space. All input variables are normalized to the range $[0,1]$ to eliminate scale differences across dimensions. Input sequences are randomly sampled, and the corresponding system outputs are obtained by simulating the system structure described in Equation (\ref{eq2}), with additive Gaussian noise introduced to emulate measurement uncertainty. Both the input and output spaces are three-dimensional. Notably, the simulated system outputs are used exclusively during the evaluation phase and are not accessed during the policy training process.

As summarized in Table (\ref{tab3}). For policy learning, $Dataset_1$ is used to train the PPO agent, while the $Dataset_2$ is reserved for evaluating parameter identification performance. This split is adopted to ensure sufficient data diversity for policy learning while maintaining an unbiased evaluation set. Furthermore, $Dataset_3$ is used as an independent dataset for evaluating the performance of methods.

\subsubsection{Baseline and Metrics}

\paragraph{PPO (Proposed Method)} The proposed PPO-based method employs a reinforcement learning model to autonomously select an informative subset of postures based on the current state vector.

\paragraph{Random Strategy:} In each episode, $4$ postures are randomly selected from the $50$ available postures. The process is repeated across multiple episodes, and the resulting performance metrics are averaged to establish a baseline.

\paragraph{Determinant of Information Matrix} According to Equation (\ref{eq2}), each episode select $K$ from $50$ input postures constitute matrix A, the determinant of the selected input posture matrix reflects the volume of the corresponding subspace, which serves as an indirect measure of the diversity of the selected posture subset. Under different strategies, a larger determinant indicates a richer and more geometrically well-distributed set of postures, providing better coverage of the input space and improving the parameter estimation quality in the system identification task.

\paragraph{Reward (PPO only)} For the proposed PPO-based method, the  reward is used to assess the convergence quality and overall effectiveness of the RL policy. Higher rewards indicate that the agent has successfully learned to select postures that maximize information gain or other task-specific objectives. It is worth noting that rewards function is designed based on the determinant of the information matrix of each episode. The  reward, which depends on $\det(\mathbf{S})$ and is computed once after all $K$ postures have been selected. 

\paragraph{Variance and Stability Indicators}
The standard deviation of repeated experiments is used to evaluate the stability and robustness of each algorithm. By running multiple independent trials, both the mean and variance of the determinant and reward metrics are calculated. A lower variance indicates more consistent performance across different runs and ensures the reliability of the proposed posture selection method under varying conditions.

\subsubsection{PPO Training and Policy Learning}
Following the simulation-guided reinforcement learning framework introduced in Section~X, Proximal Policy Optimization (PPO) is instantiated to learn an experimental point selection policy under a discrete action space and sparse terminal rewards.

The state representation is a 210-dimensional vector that aggregates multiple sources of information relevant to experimental point optimization. Specifically, it consists of pose-related features (150 dimensions), a binary availability indicator for candidate poses (50 dimensions), statistical summaries of observed system responses (6 dimensions), a scalar progress indicator encoding the current step within an episode (1 dimension), and auxiliary matrix-derived features (3 dimensions). This state formulation enables the policy to jointly reason about geometric coverage, feasibility constraints, and accumulated identification information.

The action space is discrete, defined over a finite set of 50 candidate poses. At each decision step, the policy selects a single pose from the currently available candidates. An episode corresponds to the sequential selection of four poses, after which the episode terminates automatically.
\begin{figure}[h!]
	\centering
	
	\includegraphics[width=\textwidth]{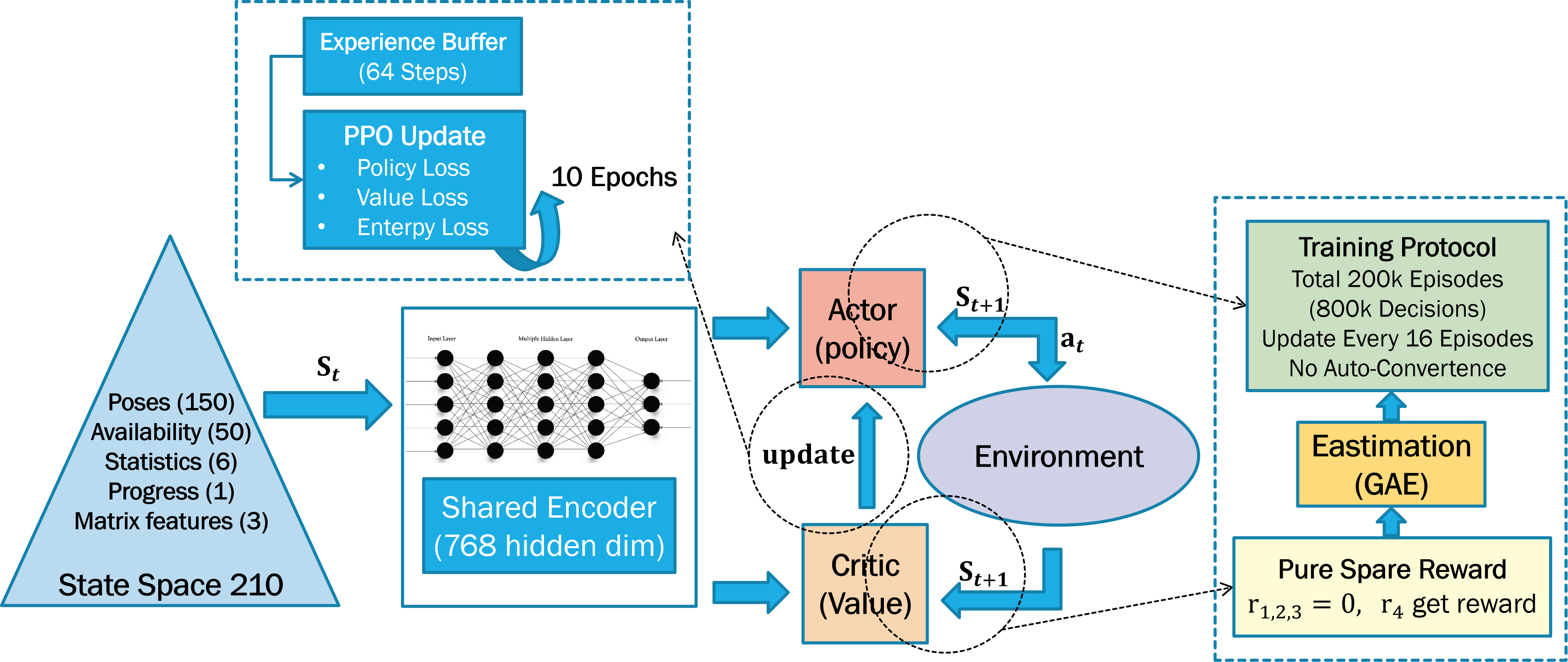}
	
	\caption{Architecture of PPO-based RL Framework for 4-Step Sequential Pose Selection.}
	\label{ppo}
\end{figure}

A sparse reward scheme is adopted to align policy learning with the objective of experimental informativeness. No intermediate rewards are provided during pose selection. Upon episode termination, a scalar reward proportional to the log-determinant of the resulting information matrix is assigned. To stabilize policy optimization under sparse terminal rewards,  scaled by a factor of 1/10 to stabilize policy updates. Under this formulation, the observed episode-level rewards typically fall within the range of 3.0 to 4.2.

The policy and value functions are parameterized by a neural network architecture with a shared encoder and separate actor and critic heads. Approximately 71\% of the network parameters are shared, promoting feature reuse between policy evaluation and value estimation. The hidden representation dimension is set to 768, resulting in a total parameter count of approximately 3.5 million. The actor outputs a categorical action distribution via a softmax layer, from which actions are sampled during training.

Training is conducted for a fixed budget of 200,000 episodes, corresponding to 800,000 decision steps in total. Policy updates are performed every 16 episodes, with each update consisting of 10 optimization epochs over the collected rollout data. A fixed training horizon is used to ensure consistency and fair comparison across different baselines under sparse-reward conditions.
\begin{table}[t]
	\centering
	\caption{PPO Training Configuration Summary}
	\label{tab2}
	\renewcommand{\arraystretch}{1.1}
	\begin{tabular}{p{2.5cm} p{3cm} p{6cm}}
		\hline
		\textbf{Module} & \textbf{Item} & \textbf{Specification} \\
		\hline
		State Space
		& Dimension
		& 210 \\
		
		& Composition
		& Poses (150) + Availability (50) + Statistics (6) + Progress (1) + Matrix features (3) \\
		
		\hline
		Action Space
		& Type
		& Discrete \\
		
		& Action set
		& $\{0, \ldots, 49\}$ \\
		
		& Semantics
		& Select one available pose per step \\
		
		\hline
		Reward Design
		& Reward type
		& Sparse \\
		
		& Intermediate reward
		& 0 \\
		
		& Terminal reward
		& $\mathrm{logdet} / 10$ \\
		
		& Reward scale
		& $[3.0,\,4.2]$ \\
		
		\hline
		Episode \\Definition
		& Episode length
		& 4 steps \\
		
		& Episode meaning
		& Sequential selection of 4 poses \\
		
		& Termination condition
		& $\text{len(selected)} = 4$ \\
		
		\hline
		Policy Network
		& Architecture
		& Shared encoder + separate actor/critic heads \\
		
		& Parameter sharing
		& $\sim$71\% \\
		
		& Hidden dimension
		& 768 \\
		
		& Total parameters
		& $\sim$3.5M \\
		
		\hline
		Output \\Parameterization
		& Action distribution
		& Categorical \\
		
		& Implementation
		& Softmax + sampling \\
		
		\hline
		Training \\Protocol
		& Training budget
		& 200k episodes \\
		
		& Total decisions
		& 800k steps \\
		
		& Update frequency
		& Every 16 episodes \\
		
		& Optimization epochs
		& 10 per update \\
		
		\hline
		Convergence \\Criterion
		& Automatic convergence
		& Not used \\
		
		& Training termination
		& Fixed 200k episodes \\
		
		\hline
	\end{tabular}
\end{table}

\subsubsection{Simulation Results Analysis}\label{sec_sim}

\begin{figure}[h!]
	\centering
	
	\includegraphics[width=\textwidth]{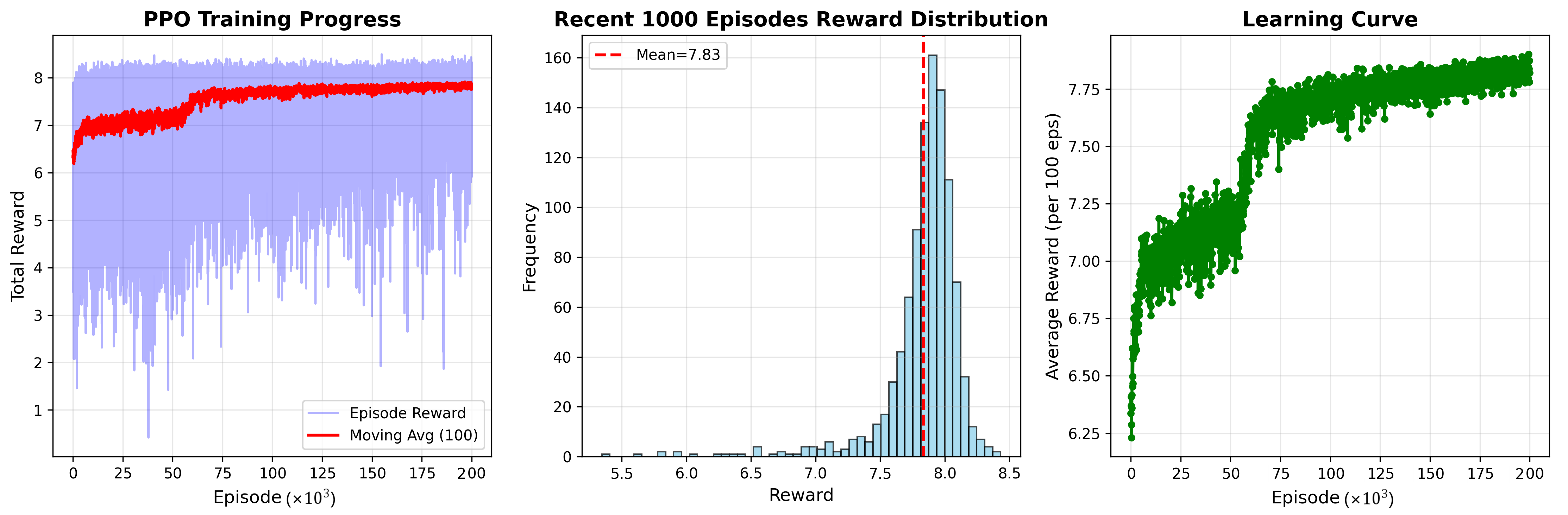}
	
	\caption{
		\textbf{a}.Training reward and moving average over episodes (left), \textbf{b}.Reward distribution over the final 1000 episodes (middle), \textbf{c}.Smoothed learning curve of average episode reward (right).}
	\label{train}
\end{figure}

\paragraph{Convergence and Learning Behavior}
Figure~\ref{train} illustrates the convergence and learning behavior of the proposed PPO-based policy. As shown in Fig.~\ref{train}a, the episode reward exhibits a clear upward trend with decreasing variance as training progresses. The smoothed learning curve in Fig.~\ref{train}c further indicates that the policy gradually converges to a stable performance plateau. To assess stability in the converged regime, Fig.~\ref{train}b reports the reward distribution over the final $1000$ episodes, which demonstrates a concentrated distribution with low variance, confirming stable policy behavior.

\paragraph{Performance Comparison with Baselines}

\begin{figure}[h!]
	\centering	
	\includegraphics[width=1\textwidth]{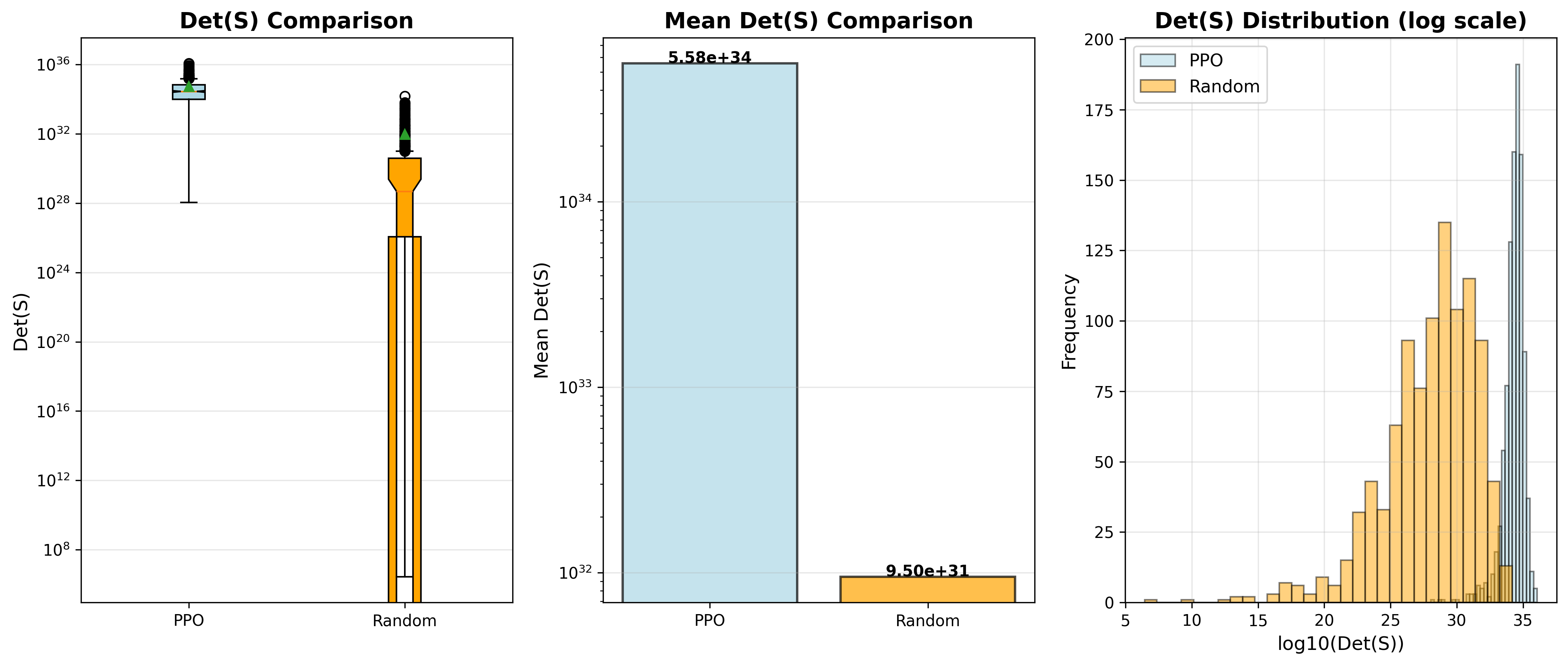}	
	\caption{
		\textbf{a}.Compare $\det(\mathbf{S})$ with baseline, 
		\textbf{b}.Compare mean $\det(\mathbf{S})$ with baseline, \textbf{c}.Compare $\det(\mathbf{S})$ distribution with baseline in log scale.}
	\label{sim_test}
\end{figure}

Figure~\ref{sim_test} compares the proposed PPO-based method with the baseline strategy, a random selection, in terms of the determinant of the information matrix. As shown in Fig.~\ref{sim_test}a, the PPO policy consistently achieves higher $\det(\mathbf{S})$ values with reduced variance, indicating both superior performance and improved stability. 
The mean $\det(\mathbf{S})$ values reported in Fig.~\ref{sim_test}b further demonstrate that PPO significantly outperforms the random strategy by more than two orders of magnitude. To provide a distribution-level comparison, Fig.~\ref{sim_test}c illustrates the log-scaled $\det(\mathbf{S})$ distributions over multiple trials. The PPO distribution is clearly right-shifted and more concentrated, confirming that the observed performance gains are consistent rather than arising from occasional high-reward samples.

\paragraph{Generalization under Independent Simulation Runs}

\begin{figure}[h!]
	\centering
	
	\includegraphics[width=1\textwidth]{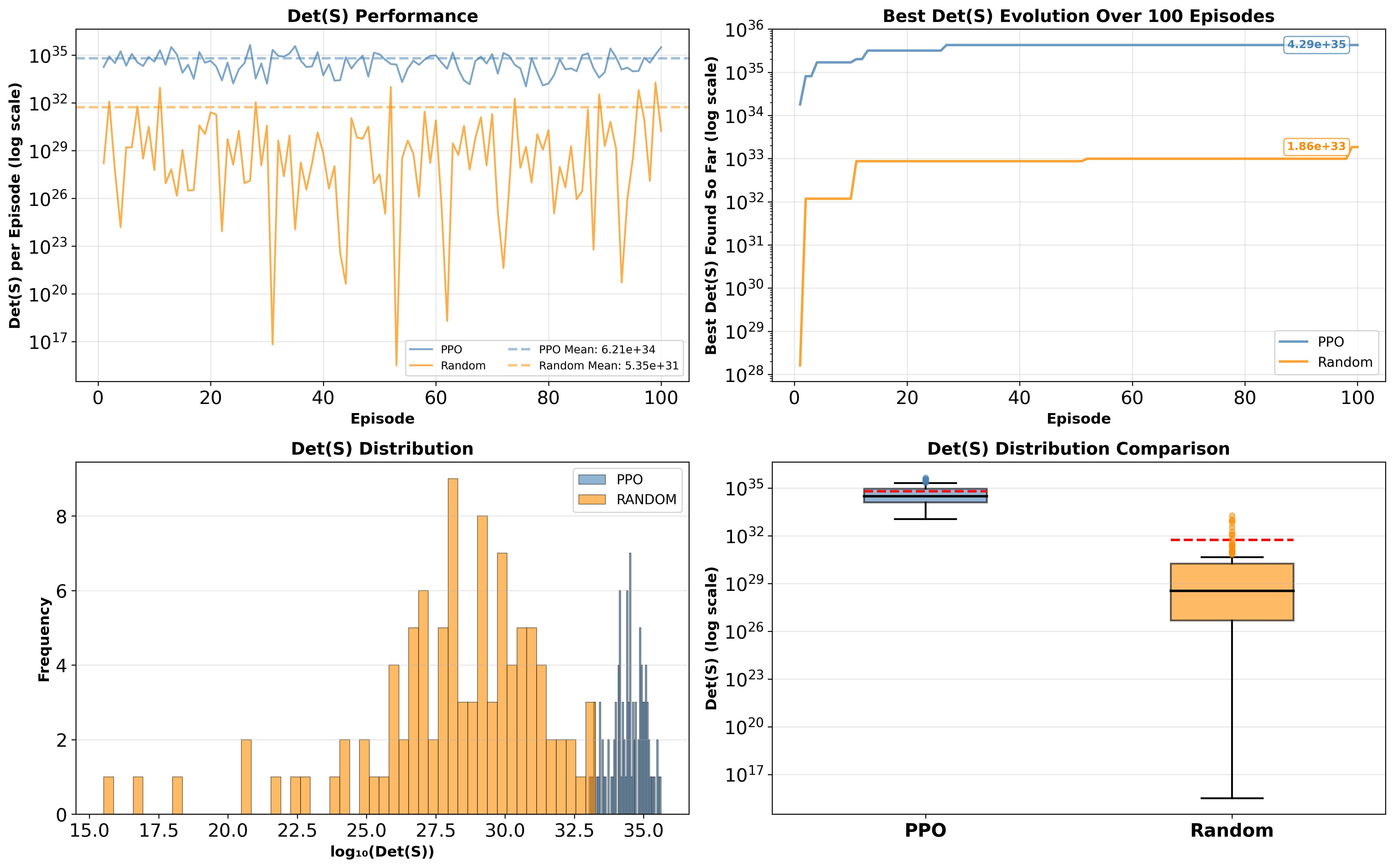}
	
	\caption{Independent Simulation Validation (performance comparison on an $Dataset_3$ generated from the same underlying distribution as the training environment, the evaluated data were not used during PPO training).}
	\label{best_det_test}
\end{figure}

To further assess the robustness of the learned policy, we evaluate PPO on an independent simulation dataset ($Dataset_3$, see Table \ref{tab3}) generated from the same underlying distribution as the training environment . Importantly, $Dataset_3$ is only used for evaluation and does not contribute to policy updates.
As shown in Fig.~\ref{best_det_test}, PPO maintains a consistently high $\det(\mathbf{S})$ value across episodes, significantly outperforming the random baseline in terms of both average performance and stability. The best $\det(\mathbf{S})$ found by PPO rapidly approaches the upper range of the distribution, whereas the random strategy exhibits slow improvement and high variance. These results indicate that the learned policy generalizes well beyond the specific simulation trajectories encountered during training.

\paragraph{Parameter Estimation Stability and Output Diversity Analysis}

\begin{figure}[h!]
	\centering
	
	\includegraphics[width=1\textwidth]{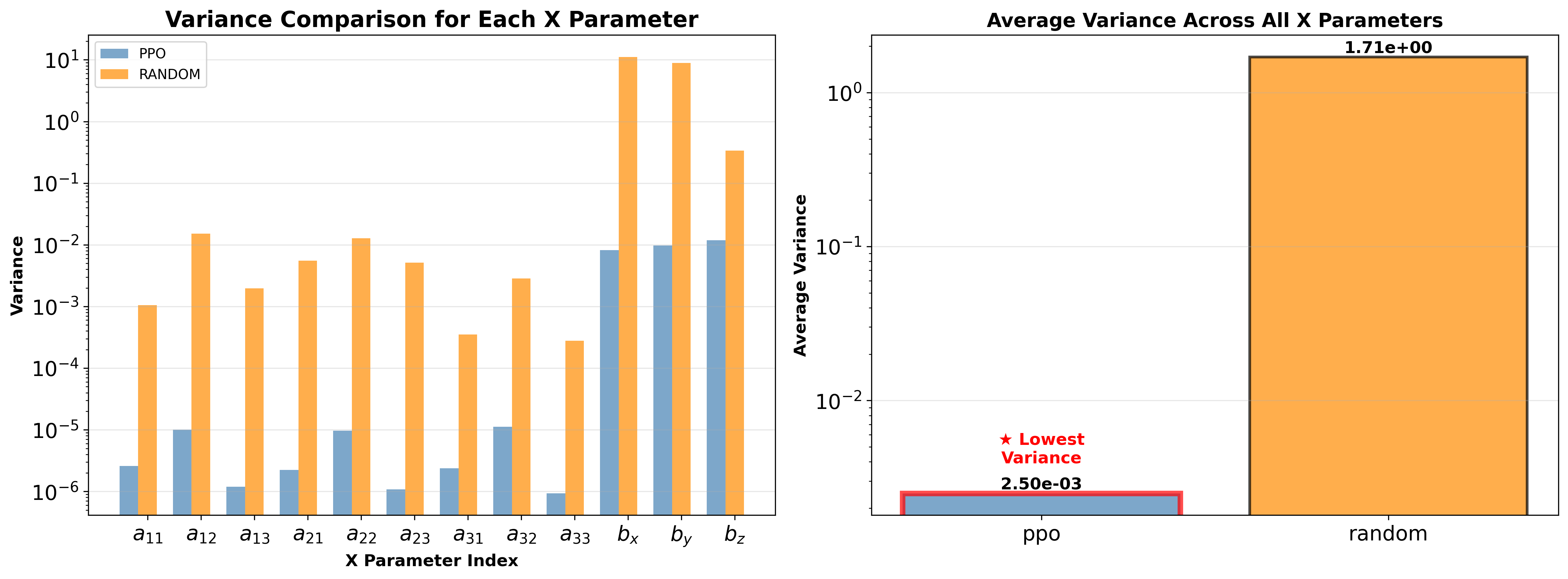}
	
	\caption{Cross-episode variance of the estimated parameters evaluated on the $Dataset_3$ under different experimental point selection strategies.}
	\label{est_X}
\end{figure}

Using $Dataset_3$, we evaluate the cross-episode stability of parameter estimation ($\mathbf{X}$) with different point selection strategies.
The variance of the estimated parameters across episodes directly reflects the stability and consistency of the identification results. Figure~\ref{est_X} shows that PPO achieves the lowest cross-episode parameter variance among all compared methods, demonstrating its ability to select informative experimental points that lead to robust parameter estimation.

\section{Experimental Validation on the Physical Robot}
\begin{figure}[h!]
	\centering
	\vspace{-5em}
	\includegraphics[width=0.8\textwidth]{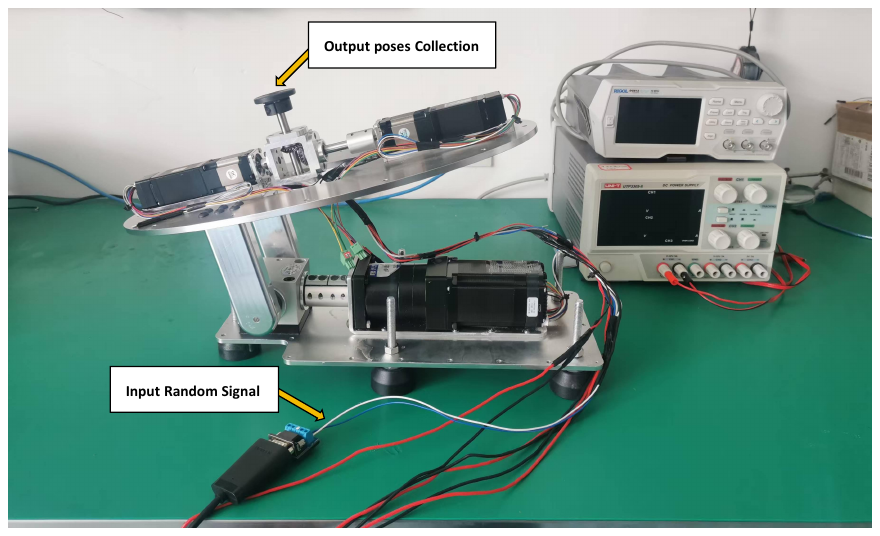}
	\vspace{-35em}
	\caption{Experimental Platform.}
	\label{fig_8}
\end{figure}
In this section, real-world experiments are divided into two stages. The first part validates the effectiveness of the Kronecker-product-based parameter identification method in a real-world environment. The second part analyzes the performance of the proposed PPO-based strategy when applied to experimental point selection.

\subsection{Real-World Experimental Setup}
\paragraph{Hardware platform} 
We utilize a 3-DOF ankle rehabilitation device (Fig.~\ref{fig_8}), controlled by three motors that facilitate both independent and coupled movements. 

\paragraph{Sensors and sampling}
A smartphone which is equipped with an IMU solution application is mounted on the end-effector, providing real-time attitude feedback for pitch, roll, and yaw axes. The IMU solution application communicates with the computer via the TCP/IP protocol, while the device’s motors communicate using the 485/Modbus-RTU protocol (Fig.~\ref{fig_8}).
 
\subsection{Real-time Data collection}

When the IMU mounted on the end-effector measures the attitude as ($0.00$, $0.00$, $0.00$), this is defined as the initial posture of the ankle rehabilitation device and serves as the zero reference point for all motor movements.

From this initial posture, random values within the pulse count range for the ankle's motion are generated. The motors then guide the rehabilitation device’s end-effector to achieve the specified posture, recording the motor pulse counts $O$($o_x$,$o_y$,$o_z$) for pitch, yaw, and roll, along with the corresponding IMU feedback posture angles $Y$($y_x$, $y_y$, $y_z$).

To ensure the accuracy of the collected postures, data acquisition is performed only after the device has fully settled at each new posture. This stabilization requirement significantly increases the time cost of the experimental procedure.

The relationship between motor pulse count and rotation angle is:
\begin{equation} \label{eq7}
	\theta =\frac{360 \, p}{\omega \,\kappa}.
\end{equation}

In Equation (\ref{eq7}), the parameter $p$ represents the motor pulse count, where the sign of $p$ indicates the direction of motor rotation. The parameter $\omega$ stands for pulses per revolution, and $\kappa$ is the reduction ratio. Based on Equation (\ref{eq7}), the pulse counts $I$($i_x$,$i_y$,$i_z$) are converted to the posture angles $U$($u_x$,$u_y$,$u_z$).

According to Table (\ref{tab3}), real-time totally contains 550 pairs of system input postures $U$($u_x$,$u_y$,$u_z$) and the corresponding system output postures $Y$($y_x$,$y_y$,$y_z$).

\subsection{Real-World Experiment Results and Analysis}
Four real-world experiments are conducted in this part to comprehensively evaluate the proposed methods.
First, the effectiveness of the Kronecker-product-based parameter identification method is validated under real experimental conditions.
Second, the PPO model with frozen weights is tested on an independent real-world dataset, and its performance is compared with a random selection baseline to assess the effectiveness of the proposed experimental point selection strategy.
Third, using the same dataset, the system parameter vector $X$, consisting of 12 unknown parameters, is estimated to evaluate identification accuracy.
Finally, the variance of the output posture vector 
$Y$, which includes three dimensions, is analyzed to assess estimation stability.
All experiments are conducted using real-time experimental data, thereby validating the practical applicability and effectiveness of the proposed framework in real-world operational settings.
\subsubsection{Real-World Experimental Evaluation of the Kronecker-Product-Based Calibration Method}
\begin{figure}[h!]
	\centering
	\includegraphics[width=0.8\textwidth]{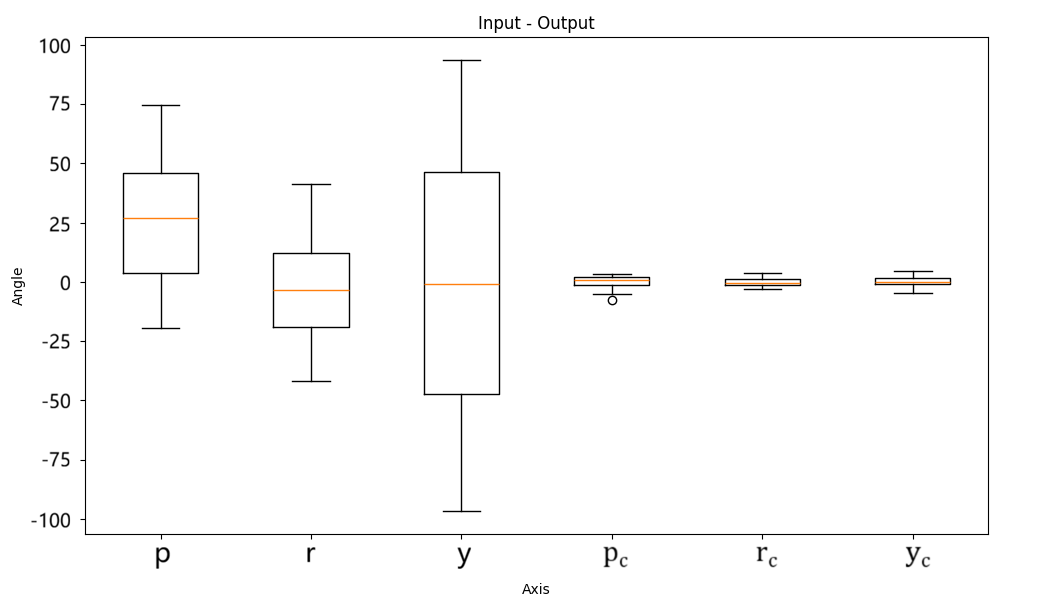}
	\caption{Distribution of input-output errors before and after open-loop calibration in real-world experiments.}
	\label{fig:ex}
\end{figure}
Figure~\ref{fig:ex} presents the statistical distributions of the posture errors before and after calibration obtained from the real-world experimental dataset. Specifically, $(p,r,y)$ denote the posture discrepancies between the motor input and the end-effector output before compensation, while $(p_c,r_c,y_c)$ represent the corresponding posture discrepancies after applying the proposed calibration method.
	
As shown in the Fig.~\ref{fig:ex}, the uncompensated posture errors exhibit large dispersion, particularly along the yaw axis, indicating significant misalignment and coupling effects in the raw system. After compensation, the posture errors in all three axes are substantially reduced and tightly clustered around zero, demonstrating that the proposed Kronecker-product-based open-loop calibration effectively corrects the input–output misalignment.

Moreover, the error reduction trends observed in the real-world experiments are consistent with those obtained in the simulation results (Fig.~\ref{fig:0}) presented in Section~\ref{Simulation Results and Analysis}, confirming the robustness and practical applicability of the proposed calibration method under real operating conditions.
\subsubsection{RL-based DoE Performance compared with Baseline on Real-time Dataset}
\begin{figure}[h!]
	\centering
	
	\includegraphics[width=1\textwidth]{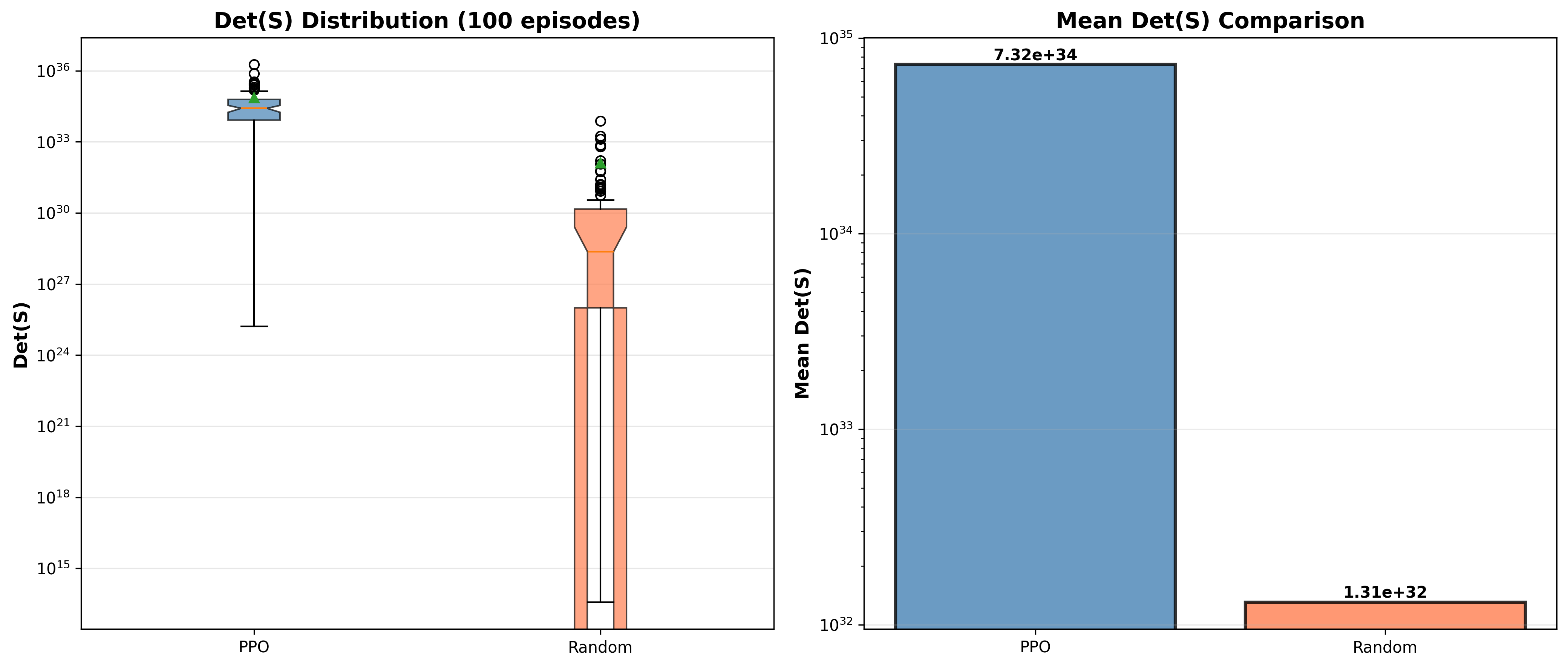}
	
	\caption{Test on independent real-time dataset.}
	\label{real_test}
\end{figure}
In section \ref*{sec_sim}, the simulation results shows that the PPO strategy compared with a random baseline at experimental points selection has a better performance. Furthermore, in order to exam different strategies' performance on the real-time dataset, we use an independent real-time dataset content 100 episode ($Dataset_4$, shown in Table \ref{tab3}). Figure~\ref*{real_test} presents the comparison of $\det(\mathbf{S})$ achieved by PPO, and Random strategies. The determinant of the state-related matrix $\mathbf{S}$ is used as an indicator of the informativeness of the selected experimental configurations.

PPO achieves the highest mean $\det(\mathbf{S})$, with values consistently concentrated in the upper range, indicating that the selected experimental points provide highly informative system observations. It is worth noting that under fixed PPO configs and dataset, the random strategy still exhibits significantly lower median values and a wide dispersion with sporadic outliers, suggesting unstable and inconsistent information acquisition.

\subsubsection{Efficiency and Robustness of RL-based DoE}
\begin{figure}[h!]
	\centering
	
	\includegraphics[width=1\textwidth]{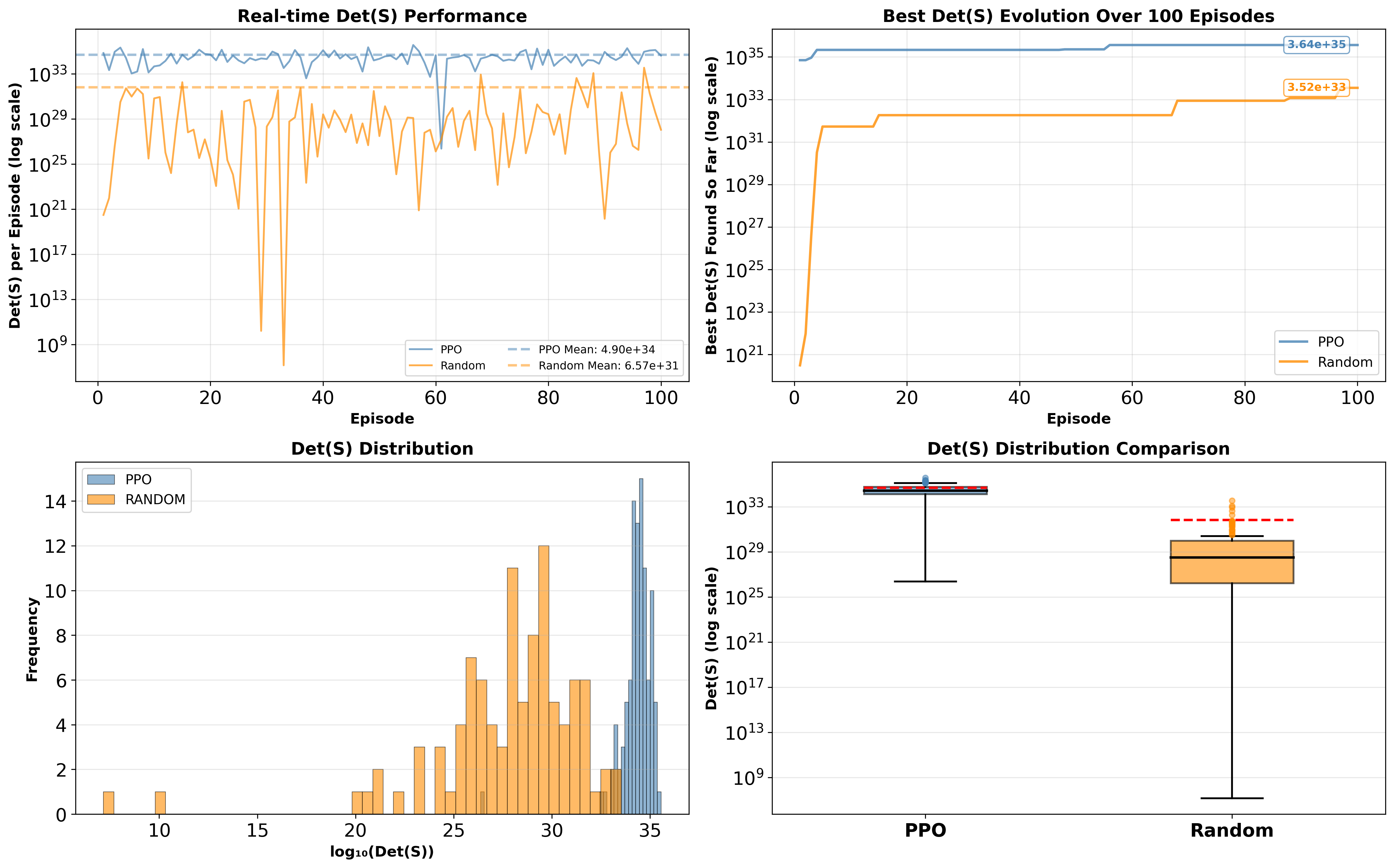}
	
	\caption{Real-world verification.}
	\label{best_det_test_real}
\end{figure}

According to Fig.~\ref{best_det_test_real}, the performance of different methods in maximizing $\det(\mathbf{S})$ is evaluated on the same real-time dataset ($Dataset_4$). PPO maintains consistently high $\det(\mathbf{S})$ values across episodes, significantly outperforming the random baseline in terms of both average performance and stability. The maximum $\det(\mathbf{S})$ achieved by PPO rapidly approaches the upper range of the observed distribution, whereas the random strategy exhibits slower improvement and substantially higher variance. These results demonstrate the robustness and effectiveness of PPO for experimental point selection on real-time data.

\begin{figure}[h!]
	\centering
	
	\includegraphics[width=1\textwidth]{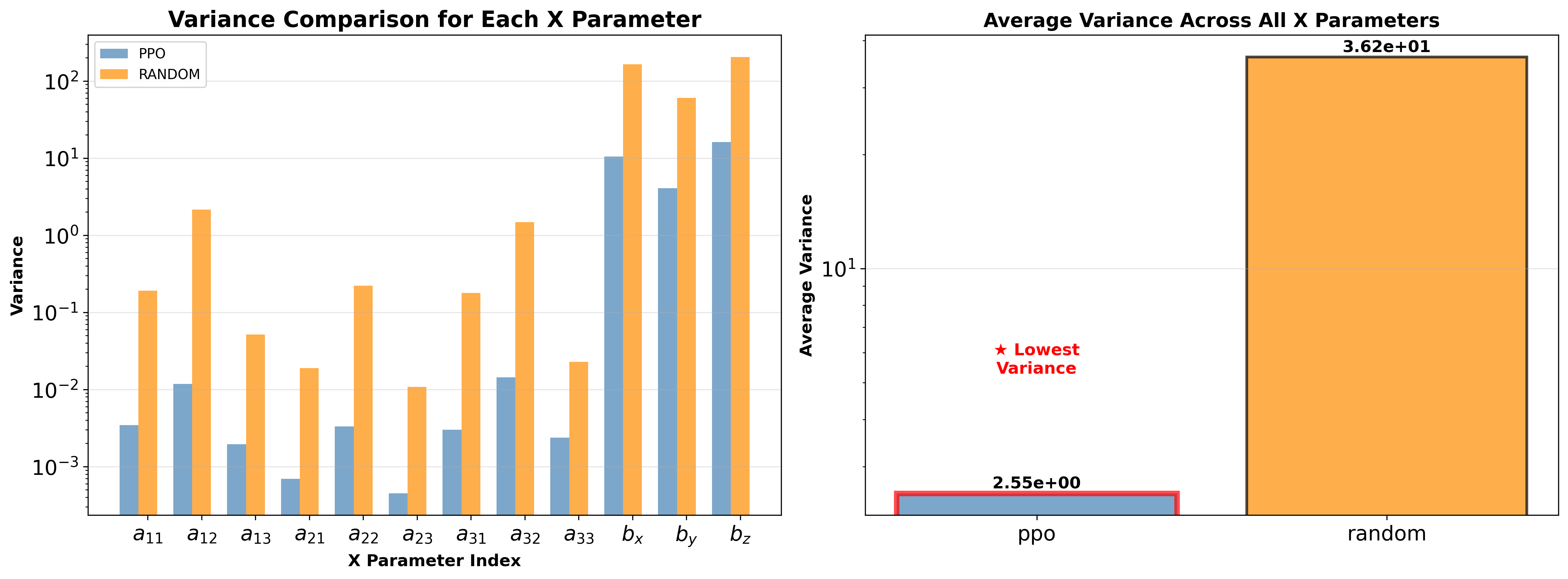}
	
	\caption{Identify parameter vector X variance across 100 episodes from real-time datasets.}
	\label{est_X_real}
\end{figure}

Figure~\ref*{est_X_real} evaluates the variance of the estimated system parameters across episodes on $Dataset_4$ (shown in Table \ref{tab3}), which directly reflects the stability and consistency of parameter identification results. A total of 12 parameters are analyzed, and significant differences in estimation stability are observed among the two strategies.

PPO consistently achieves the lowest parameter variance across most parameters, indicating robust and repeatable identification results, whereas the Random strategy results in extremely large parameter variance, spanning several orders of magnitude for certain parameters. These results demonstrate that random experimental selection fails to provide reliable information for stable parameter estimation.

\subsubsection{Parameter Identification Performance with and without RL-based DoE}	

\begin{figure}[h!]
	\centering
	
	\includegraphics[width=0.9\textwidth]{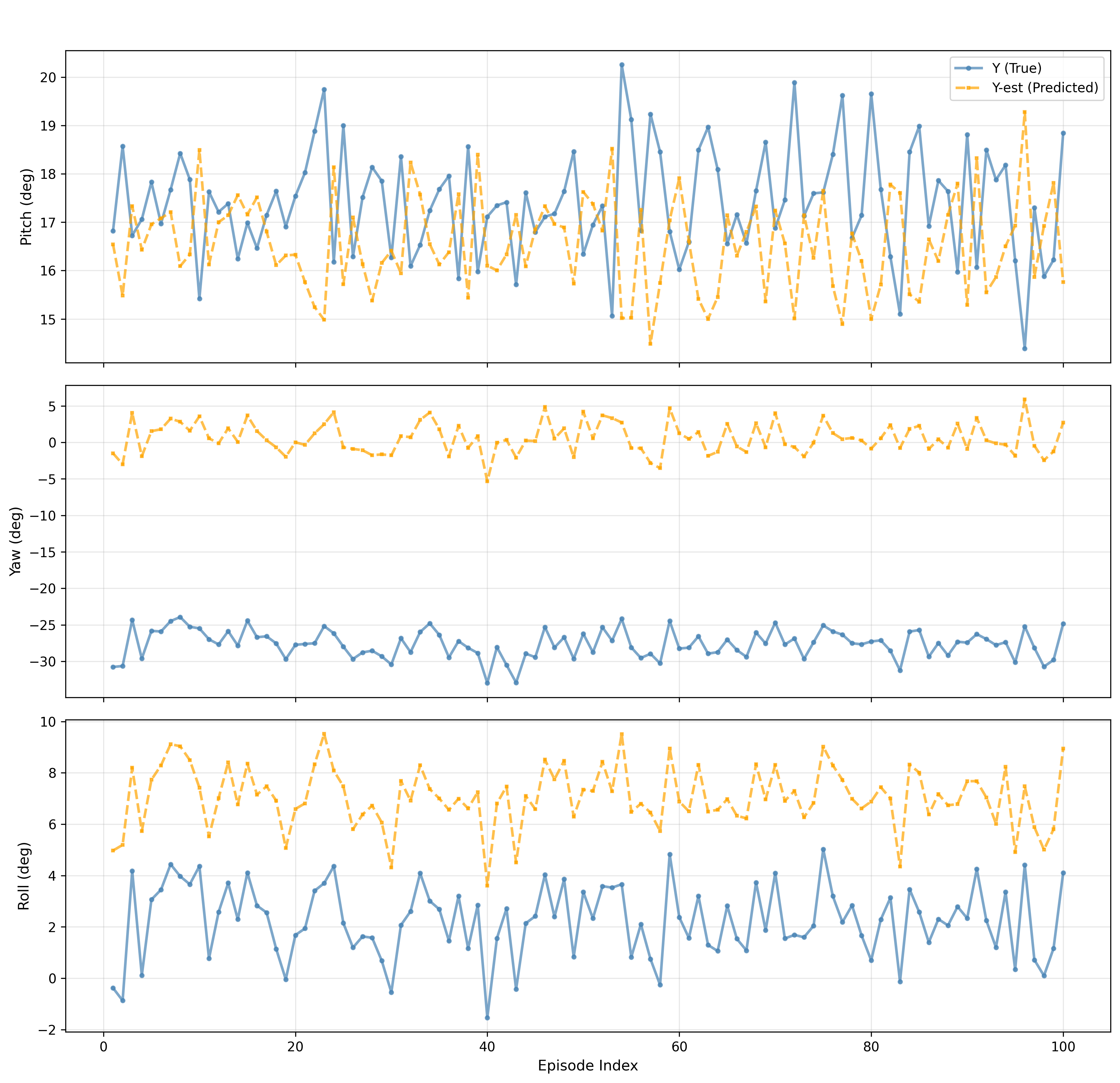}
	
	\caption{The estimated X from the random combination of previous real datasets is applied to estimate eight points of Y across other different episodes of real datasets.}
	\label{best_X0_est_8Y}
\end{figure}

\begin{figure}[h!]
	\centering
	
	\includegraphics[width=0.9\textwidth]{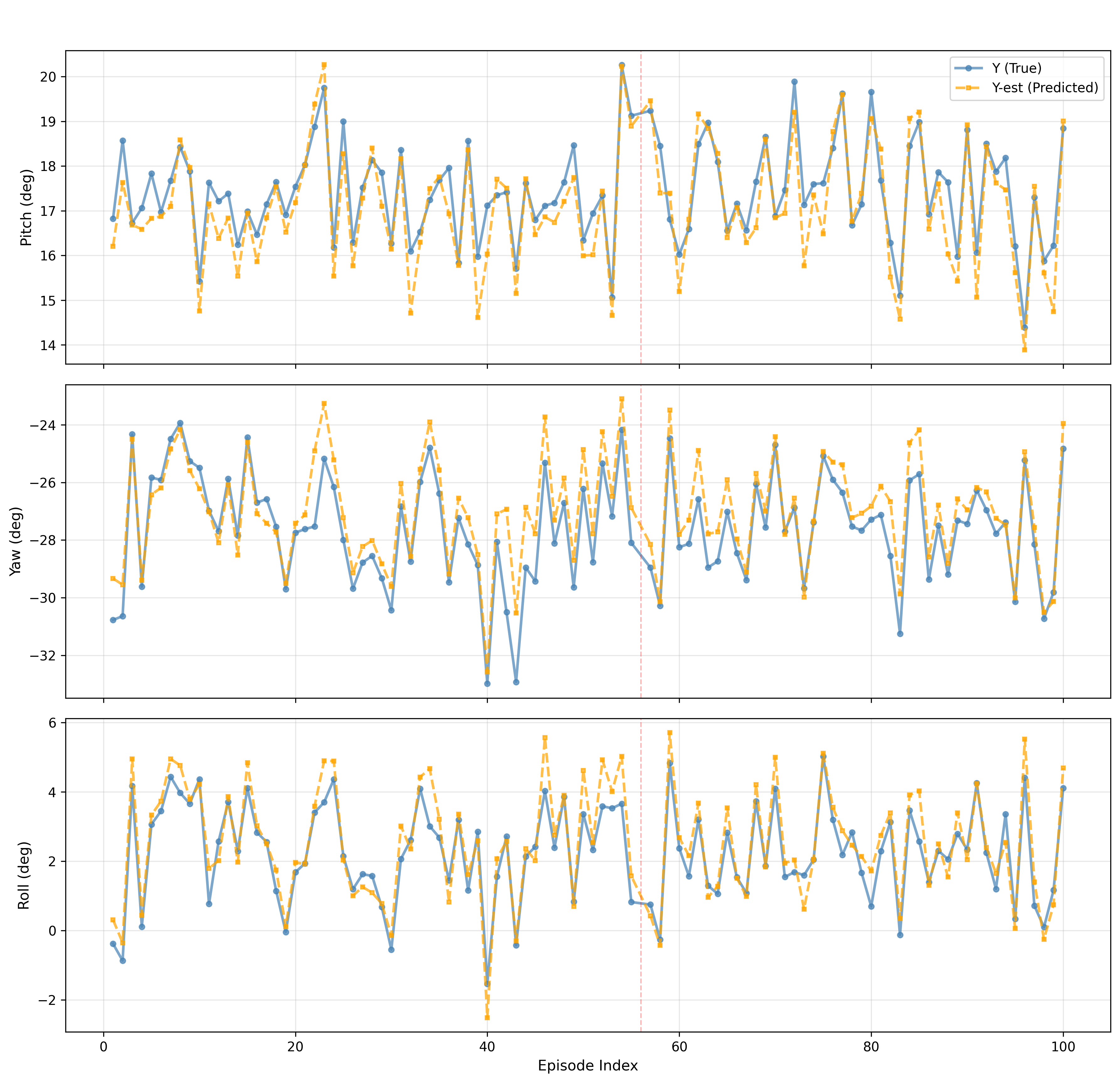}
	
	\caption{The estimated X from the best deterministic combination of real datasets is applied to estimate eight points of Y across other different episodes.}
	\label{best_X*_est_8Y}
\end{figure}

A clear performance difference can be observed by comparing Fig.~\ref{best_X0_est_8Y} and Fig.~\ref{best_X*_est_8Y} corresponding to the parameter vectors $\mathbf{X_0}$ and $\mathbf{X_*}$, respectively.
$\mathbf{X_0}$ is estimated from four informative postures selected by the proposed PPO-based method from an earlier episode of the real-world experimental dataset, whereas $\mathbf{X_*}$ is estimated from a posture combination that maximizes $\det(\mathbf{S})$ based on a later episode.

Figure~\ref{best_X0_est_8Y} illustrates the predicted outputs obtained using $\mathbf{X_0}$ (predict eight points from each episode), when applied across episodes, noticeable discrepancies between the predicted outputs $\mathbf{Y_{est}}$ and the true outputs $\mathbf{Y}$ can be observed, particularly in the Yaw dimension, the predicted signal exhibits a pronounced systematic bias, indicating limited generalization capability of $\mathbf{X_0}$ when evaluated beyond the episode from which it was identified.

In contrast, Fig.~\ref{best_X*_est_8Y} presents the prediction results obtained using $\mathbf{X}_*$,  although this parameter vector is identified using only four carefully selected postures within a single episode, the resulting predictions show a substantially higher level of agreement with the true outputs across all three output dimensions. The predicted curves closely follow the true signals in both trend and magnitude, demonstrating improved consistency and reduced estimation error across episodes.

This comparison clearly indicates that the parameter vector $\mathbf{X_*}$, derived from a later episode, achieves superior prediction accuracy and stronger cross-episode generalization than $\mathbf{X_0}$, which is estimated from an earlier episode, when evaluated on the same real-time dataset. These results highlight the effectiveness of maximizing $\det(\mathbf{S})$ as a criterion for experimental point selection, as well as the necessity of periodic calibration, enabling reliable parameter identification with fewer measurements.
\section{Conclusion} 
This paper first proposed a Kronecker-product-based open-loop calibration method that formulates the input-output alignment as a linear parameter identification task. It then investigates the problem of experimental point selection guided by information maximization for system parameter identification under limited experimental conditions. Based on the established Kronecker-product-based open-loop parameter identification formulation, a simulation-guided reinforcement learning framework using Proximal Policy Optimization (PPO) is proposed to automatically select informative postures that maximize the information content of the collected data, quantified by the determinant of the information matrix, $\det(\mathbf{S})$.

Extensive simulation and real-world experiments demonstrate that the proposed experimental selection method consistently outperforms random baseline strategies. The PPO-based policy successfully identifies posture combinations that yield significantly higher $\det(\mathbf{S})$ values, leading to more robust and stable parameter estimation across episodes. In particular, the learned policy achieves low cross-episode variance in the estimated parameters while avoiding excessive or unstructured variability in the identification results, indicating an effective balance between exploration and estimation stability.

Furthermore, experimental results on independent real-world datasets demonstrate that parameter vectors estimated from PPO-selected posture combinations exhibit strong generalization capability beyond the specific episodes used for identification. Despite relying on only a limited number of selected postures, the resulting parameter estimates yield accurate output predictions across different episodes and datasets, outperforming estimates obtained from larger but unstructured measurement sets. These findings confirm that maximizing $\det(\mathbf{S})$ is a meaningful and effective criterion for experimental design, and that reinforcement learning provides a powerful mechanism for discovering highly informative experimental configurations.

In summary, this study presents a practical and general framework for improving system parameter identification through learning-based experimental point optimization. By integrating a Kronecker-based open-loop identification method with a D-optimality-guided reinforcement learning strategy, the proposed approach enhances parameter observability while substantially reducing experimental burden. Importantly, the optimization process relies solely on system input information and does not require explicit analytical system models, making it particularly suitable for real-world systems with limited sensing capabilities. Overall, the proposed framework provides a general and scalable solution for experimental design and parameter identification in complex robotic and mechatronic systems.

\bibliographystyle{cas-model2-names}
\bibliography{reference}
\end{document}